\documentclass[letterpaper, 10 pt, conference]{ieeeconf}  

\IEEEoverridecommandlockouts                              

\usepackage{graphicx} 
\usepackage{dblfloatfix} 
\usepackage{amsmath} 
\usepackage{amssymb}  
\usepackage{bm}
\usepackage{multirow}

\DeclareMathAlphabet{\mbf}{OT1}{ptm}{b}{n}
\def\Vec#1{\!\!\hbox{$#1$\kern-0.38em\lower0.85em\hbox{$\vec{}\,$}}\,}
\newcommand{\bcf}{\;\mbox{\boldmath ${\cal F}$\unboldmath}}
\newcommand\norm[1]{\left\lVert#1\right\rVert}

\graphicspath{{figs/}}   

\title{\LARGE \bf
Experimental Comparison of Visual and Single-Receiver GPS Odometry
}

\author{Benjamin Congram and Timothy D. Barfoot
\thanks{Both authors are with the University of Toronto Institute
for Aerospace Studies (UTIAS), 4925 Dufferin St, Ontario,
Canada.
        {\tt\small ben.congram@robotics.utias.utoronto.ca, tim.barfoot@utoronto.ca}}%
}

\begin{document}

\maketitle
\thispagestyle{empty}
\pagestyle{empty}

\begin{abstract}

Mobile robots rely on odometry to navigate through areas where localization fails.
Visual odometry (VO) is a common solution for obtaining robust and consistent relative motion estimates of the vehicle frame.
Contrarily, Global Positioning System (GPS) measurements are typically used for absolute positioning and localization.
However, when the constraint on absolute accuracy is relaxed, time-differenced carrier phase (TDCP) measurements can be used to find accurate relative position estimates with one single-frequency GPS receiver.
This suggests practitioners may want to consider GPS odometry as an alternative or in combination with VO\@.
We describe a robust method for single-receiver GPS odometry on an unmanned ground vehicle (UGV).
We then present an experimental comparison of the two strategies on the same test trajectories.
After 1.8km of testing, the results show our GPS odometry method has a 75\% lower drift rate than a proven stereo VO method while maintaining a smooth error signal despite varying satellite availability.

\end{abstract}

\section{Introduction}\label{sec:introduction}

Odometry is an important component of almost any mobile robotic navigation strategy; it takes many forms including visual, visual-inertial, lidar, and wheel odometry.
All of these use different sensors to accomplish the common goal of estimating the vehicle's path or trajectory.
In mapping, odometry allows local reconstruction of the environment and in localization it is critical to the success of autonomous navigation systems such as Visual Teach and Repeat (VT\&R)~\cite{Furgale2010}.
In experience-based navigation (EBN)~\cite{Churchill2013} and multi-experience localization (MEL)~\cite{Paton2016}, odometry is used to bound pose uncertainty in short sections (i.e., less than 50m) where localization fails due to factors such as appearance change.
If odometry drift becomes too large, the robot may not be able to navigate safely.
This is in turn causes a missed opportunity to improve the map.
Better odometry allows a robot to dead-reckon for longer sections and therefore drive further successfully.

\begin{figure}[t]
	\centering
	\includegraphics[trim={0 0 2mm 0},clip,width=1.0\linewidth]{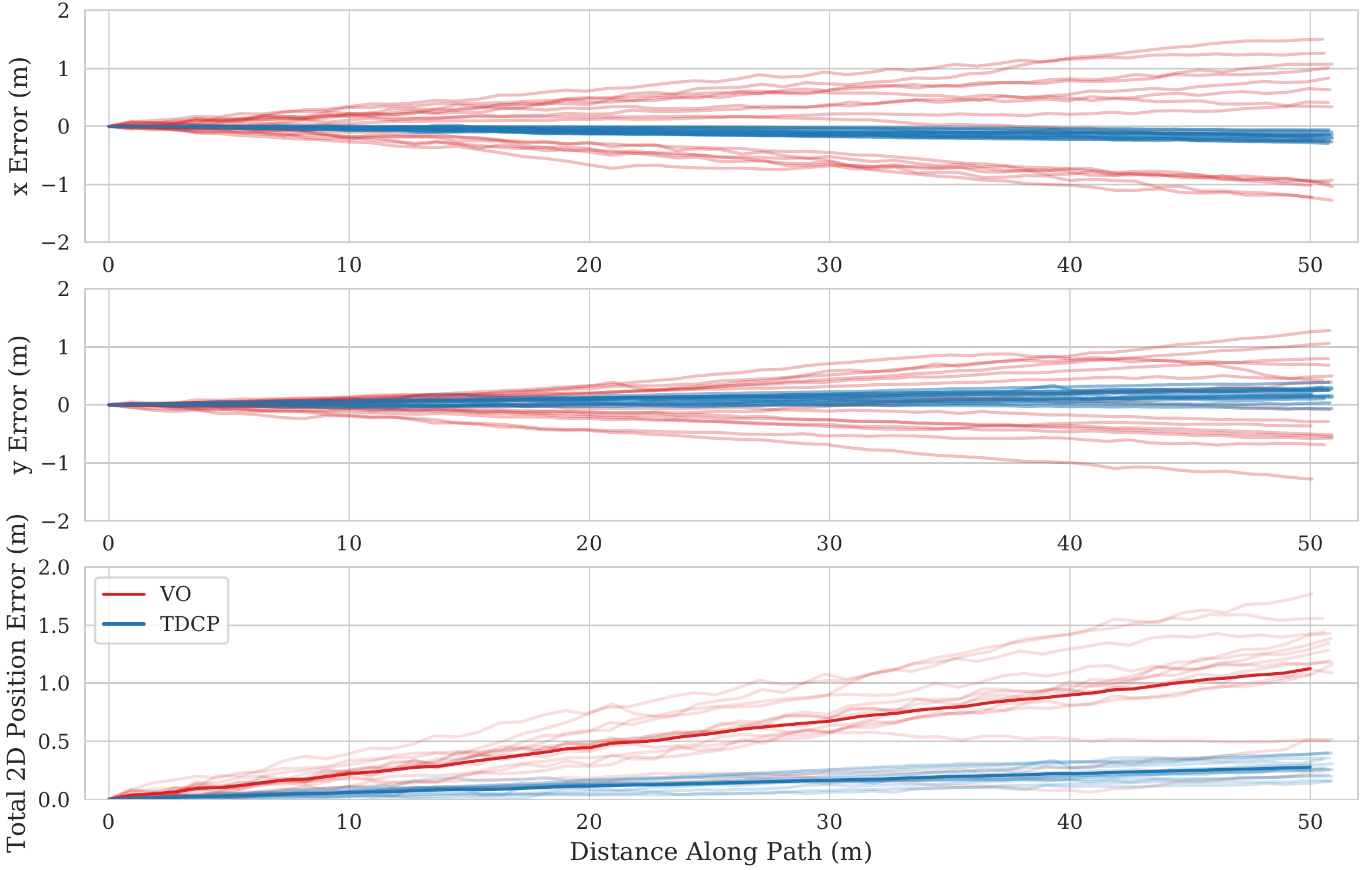}
	\caption{\footnotesize Comparison of VO and TDCP-based single-receiver GPS odometry position drift. The fainter lines represent individual trajectories while the darker line plots the average error for the algorithm.}
	\label{fig:vo-gps-errors}
\end{figure}

One idea to improve a robot's odometry is to consider other sensors.
Single-frequency Global Positioning System (GPS) receivers are now ubiquitous, coming standard in almost every smartphone.
First operational in 1983~\cite{seeber2008satellite}, GPS allows an absolute positioning solution to be calculated anywhere on Earth with a clear view of the sky.
Since then, other Global Navigation Satellite Systems (GNSS) constellations such as GLONASS, Galileo, and BeiDou have come online and may be used independently or in combination with GPS\@.
GNSS has become an important tool for mobile robotic navigation.
However, standard pseudorange GNSS positioning does not have the accuracy required to bound vehicle travel within the envelope required for visual localization, which typically degenerates with decimetre-level lateral errors~\cite{Furgale2010}.

\begin{figure}[tb]
	\centering
	\vspace{3mm}
	\includegraphics[trim={0 1.5mm 0 3.5mm},clip,width=1.0\linewidth]{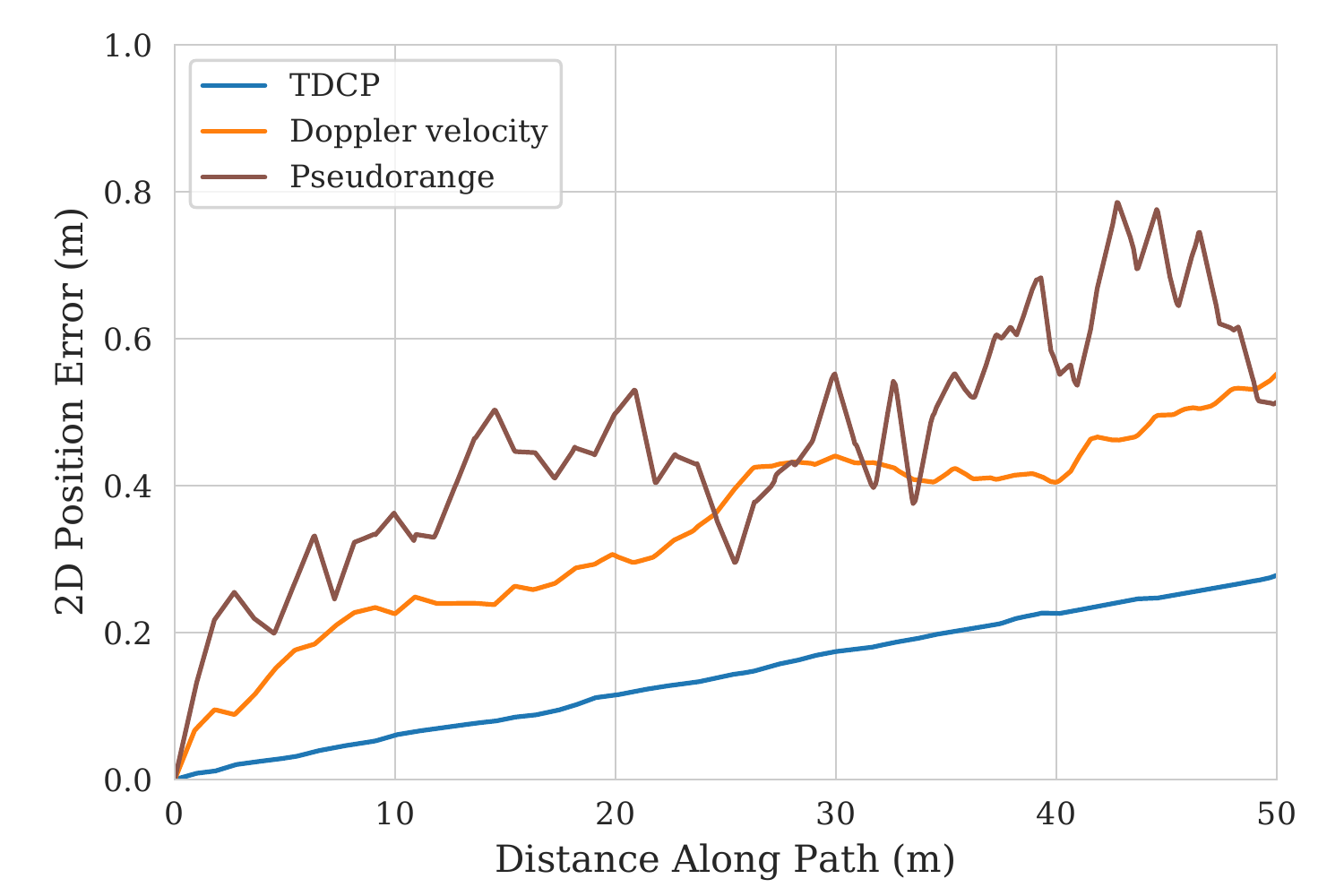}
	\caption{\footnotesize Comparison of the mean relative position error for three different techniques using a single GPS receiver across 12 independent paths. The TDCP method is both more accurate and much smoother than the pseudorange positioning while also outperforming the integrated Doppler velocity.}
	\label{fig:single-receiver-comparison}
\end{figure}

Utilizing other GNSS observables over short windows of time can improve relative positioning.
Figure~\ref{fig:single-receiver-comparison} illustrates the relative accuracy of three different single-receiver odometry strategies over a short trajectory.
In this paper, we demonstrate the use of time-differenced carrier phase (TDCP) measurements as an odometry solution for an unmanned ground vehicle (UGV).

We compare the performance of single-receiver GPS odometry with stereo visual odometry (VO) on the same set of test trajectories.
To the best of our knowledge this is the first study comparing and contrasting TDCP with VO\@.
Our results, previewed in Figure~\ref{fig:vo-gps-errors} and discussed in Section~\ref{sec:results}, show TDCP is a worthy alternative to VO in outdoor applications.
We also briefly examine the advantages of using both sensors in the same estimator in Section~\ref{subsec:combining}.

\section{Related Work}\label{sec:related-work}

\subsection{Global Navigation Satellite Systems}\label{subsec:global-navigation-satellite-systems}

Standard GNSS positioning involves the trilateration of pseudorange measurements of four or more dedicated positioning satellites.
In good conditions, pseudorange-based positioning can achieve accuracy on the order of 1--2m despite several sources of error affecting the signal's measured time-of-flight~\cite{kaplan2005understanding}.
GNSS receivers can also calculate the carrier phase of the signal based on its Doppler shift.
These measurements are much less noisy but cannot be used directly due to the unknown integer ambiguity of wavelengths between the receiver and each satellite.
Real-time kinetic (RTK) processing uses a second, static receiver nearby.
By comparing the pseudorange and carrier phase measurements of both receivers, the integer ambiguities can be resolved over time providing centimetre-level accuracy.
However, the requirement for a dedicated base station and communications link make RTK expensive and often impractical.

The idea of comparing carrier phase measurements from the same receiver at different times was first proposed by Ulmer et al.~\cite{ulmer1995accurate} but has received comparatively little attention in the robotics community.
The technique, commonly known as TDCP, was developed for static geomatic surveying~\cite{ulmer1995accurate}, \cite{Michaud2001}, \cite{Balard2006} but is easily extended to full trajectories.
When a receiver is in phase lock with a satellite, the ambiguity affecting carrier phase measurements is time invariant.
Within this period, differencing two phase measurements will cancel the ambiguity and avoid the need to resolve it.
Therefore, better accuracy can be achieved in estimating the relative receiver displacement between the two times, though absolute positioning accuracy still remains high~\cite{Michaud2001}.
TDCP has been used in applications as wide ranging as vehicle convoying~\cite{travis2010path}, \cite{pierce2017opportunistic} and bird-flight trajectory reconstruction~\cite{traugott2011precise}.
Success has been shown in combining TDCP with inertial navigation systems (INS)~\cite{Soon2008}, \cite{zhao2016applying}.

Doppler velocity integration is another technique that has been used for single-receiver positioning.
The Doppler velocity observable is closely related to the carrier phase observable but the latter should be preferred for position estimation as it is less noisy~\cite{wendel2003enhancement}.
Despite the complementary properties of GNSS and vision as sensors, TDCP has not been significantly utilized on vision-based robots.
We believe this to be the first investigation of stereo VO and TDCP together and no direct comparison to exist in the literature.

\subsection{Visual Odometry}\label{subsec:visual-odometry}

Visual odometry is the problem of estimating camera motion from a sequence of images in real-time.
The first implementation was developed by Moravec~\cite{moravec1980obstacle} for a Mars rover.
Since then it has evolved to become a standard component of mobile robotic navigation serving, for example, as an essential component of visual simultaneous localization and mapping (SLAM).
It may be used alone for dead-reckoning or fused with data from other sensors such as lidar, INS, or wheel odometry.
The basic pipeline involves detecting features in an image and tracking them in the sequence before using the geometry of those features to estimate viewpoint motion~\cite{scaramuzza2011visual}.
There are many variations with algorithms available using both monocular and stereo cameras and using both sparse features and dense correspondences~\cite{howard2008real}.
Recently, deep-learning approaches~\cite{wang2017deepvo}, \cite{xue2019beyond} to VO have gained interest, though feature-based methods still remain relevant.

We use an implementation of VO developed as a component of VT\&R based on parallel tracking and mapping (PTAM)~\cite{Klein2007}.
Motion estimates are computed at framerate while landmark positions are optimized after each keyframe.
It is fast and reliable with the parameters pre-tuned on data separate from that presented in this work.
While testing on nearly 10km of driving over 30 hours, MacTavish et al.~\cite{MacTavish2017a} found a 1.5\% translational drift rate during daytime conditions and a 2.4\% rate at nighttime via the use of headlights for this algorithm.

\section{Methodology}\label{sec:methodology}
\subsection{Coordinate Frames}\label{subsec:coordinate-frames}
There are five frames of interest to our optimization problem.
The global east-north-up (ENU) frame, $\Vec{\bcf}_g$, is a stationary frame tangential to the Earth at the vehicle start position.
All other frames are transient.
The vehicle frame, $\Vec{\bcf}_v$, is located at the centre of the vehicle at axle height.
All estimation is computed in $\Vec{\bcf}_v$ before being transformed to the GPS receiver frame, $\Vec{\bcf}_r$, for comparison with ground truth positions.
The origin of the satellite frame, $\Vec{\bcf}_s$, is defined at the antenna phase centre (APC) for calculating ranges.
Finally, the camera frame, $\Vec{\bcf}_c$, is located at the left camera of the stereo module.
The VO algorithm is also configured to output estimates in the vehicle frame.

\begin{figure*}[tb]
	\centering
	\vspace{2mm}
	\fbox{\includegraphics[trim={0 3mm 0 3mm},clip,width=0.99\textwidth]{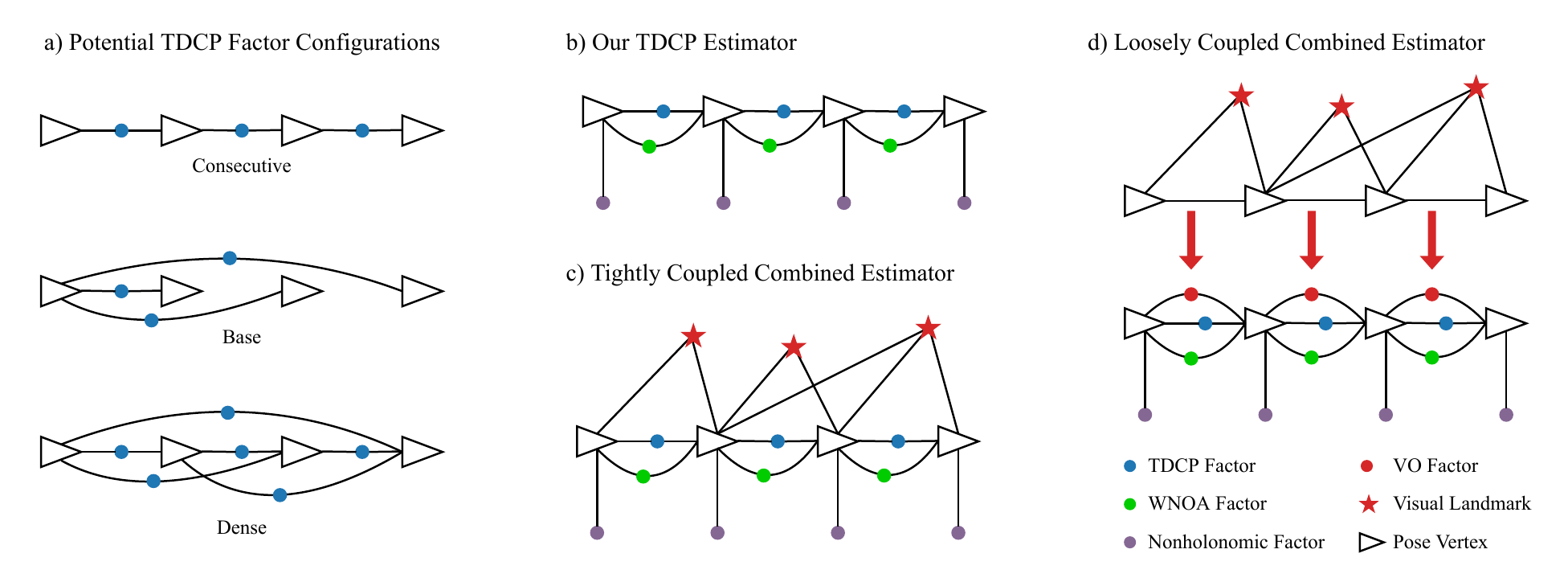}}
	\caption{\footnotesize a) Potential ways TDCP factors can be added. Due to the error characteristics of the phase range, they all give very similar position estimates. The ``Consecutive'' configuration was chosen for our estimator. b) Factor graph for our TDCP algorithm. c) Factor graph that comes from combing VO and TDCP in a tightly-coupled fashion. d) Factor graph for the loosely-coupled estimator used in Section~\ref{subsec:combining} for ease of comparison. VO is first run to estimate pose changes then those estimates are added as factors.}
	\label{fig:factor_graph}
\end{figure*}

\subsection{Time-Differenced Carrier Phase}\label{subsec:time-differenced-carrier-phase}

Whereas RTK positioning makes use of carrier phase measurements from two receivers separated in space, TDCP positioning makes use of measurements from a single receiver separated by both time and space.
The carrier phase range equation to a single satellite at time $a$ is given by
\begin{equation}
\Phi_{a} = \rho_{a} + N + c\delta^{R}_a - c\delta^{S}_a + E_{a} + T_{a} - I_{a}  + m_{a} + \epsilon_{a},     \label{eq:phase-range-single}
\end{equation}
where $\Phi_{a}$ is the measured phase in radians multiplied by the known wavelength so that all values have units of metres.
GNSS receivers can measure the incoming phase quite accurately meaning the white noise affecting the measurement, $\epsilon$, is typically less than 2mm~\cite{kaplan2005understanding}.
However, the signal is affected by several sources of systematic error as it propagates from satellite to receiver causing the measured range, $\Phi$, to differ from the true range to the satellite, $\rho$.
These include receiver and satellite clock errors ($\delta^{R}$ and $\delta^{S}$), satellite ephemeris error ($E$), tropospheric delay ($T$), ionospheric affects ($I$), and multipath ($m$).

$N$ is the unknown wavelength ambiguity; if the receiver stays in phase lock with the satellite it is time invariant.
We can therefore eliminate it by differencing (\ref{eq:phase-range-single}) taken at two times, $a$ and $b$:
\begin{equation}
\Phi_{b} - \Phi_{a} = \rho_{ba} + c\delta^{R}_{ba} - c\delta^{S}_{ba} + E_{ba} + T_{ba} - I_{ba}  + m_{ba} + \epsilon_{ba}. \label{eq:phiBA}
\end{equation}
The subscript $ba$ denotes the difference between a quantity at time $b$ and time $a$.
The receiver clock error is typically large so must be dealt with explicitly, either by estimating it or differencing the equation again for two different satellites.
The latter gives our measurement model:
\begin{equation}
\Phi_{ba}^{\mathit{21}} = \rho_{ba}^{\mathit{21}} - c\delta^{S,\mathit{21}}_{ba} + E_{ba}^{\mathit{21}} + T_{ba}^{\mathit{21}} - I_{ba}^{\mathit{21}}  + m_{ba}^{\mathit{21}} + \epsilon_{ba}^{\mathit{21}}, \label{eq:phi21BA}
\end{equation}
where $\rho_{ba}^{\mathit{21}}$, for example, denotes the double difference $\left(\rho_{b}^{\mathit{2}} - \rho_{a}^{\mathit{2}}\right) - \left(\rho_{b}^{\mathit{1}} - \rho_{a}^{\mathit{1}}\right)$.
The ranges making up $\rho_{ba}^{\mathit{21}}$ are calculated using:
\begin{equation}
\rho_a = \norm{\mbf{r}^{sr}_g(t_{a})} = \norm{\mbf{r}^{sg}_g(t_{a}) - \mbf{r}^{rg}_g(t_{a})},	\label{eq:range-from-state}
\end{equation}
where $\mbf{r}^{sg}_g$ is the known satellite ephemeris and $\mbf{r}^{rg}_g$ is our state.
It is important to recalculate the ephemeris at each measurement time because the satellites travel at 3.9km/s.
From~\eqref{eq:phi21BA}, we can write our error term for one pair of satellites seen at one pair of positions as:
\begin{equation}
e^{\mathit{21}}_{ba} = \Phi_{ba}^{\mathit{21}} - \rho_{ba}^{\mathit{21}}.\label{eq:tdcp_error}
\end{equation}
Our weighted least squares factor given $n$ commonly seen satellites at $t_a$ and $t_b$ is
\begin{equation}
J_{ba} = \sum_{k=2}^n w_k\left(e^{\mathit{k1}}_{ba}\right)^2 \label{eq:tdcp_factor}
\end{equation}
where $w_k$ is a scalar variance parameter, which we set as a constant in our implementation though it could be tuned if more information on the measurement quality from each satellite was known.
$J_{ba}$ is symbolized as a blue dot in Figure~\ref{fig:factor_graph}.

\begin{figure}[tb]
	\centering
	\vspace{2mm}
	\includegraphics[trim={0 1.5mm 0 2.5mm},clip,width=1\linewidth]{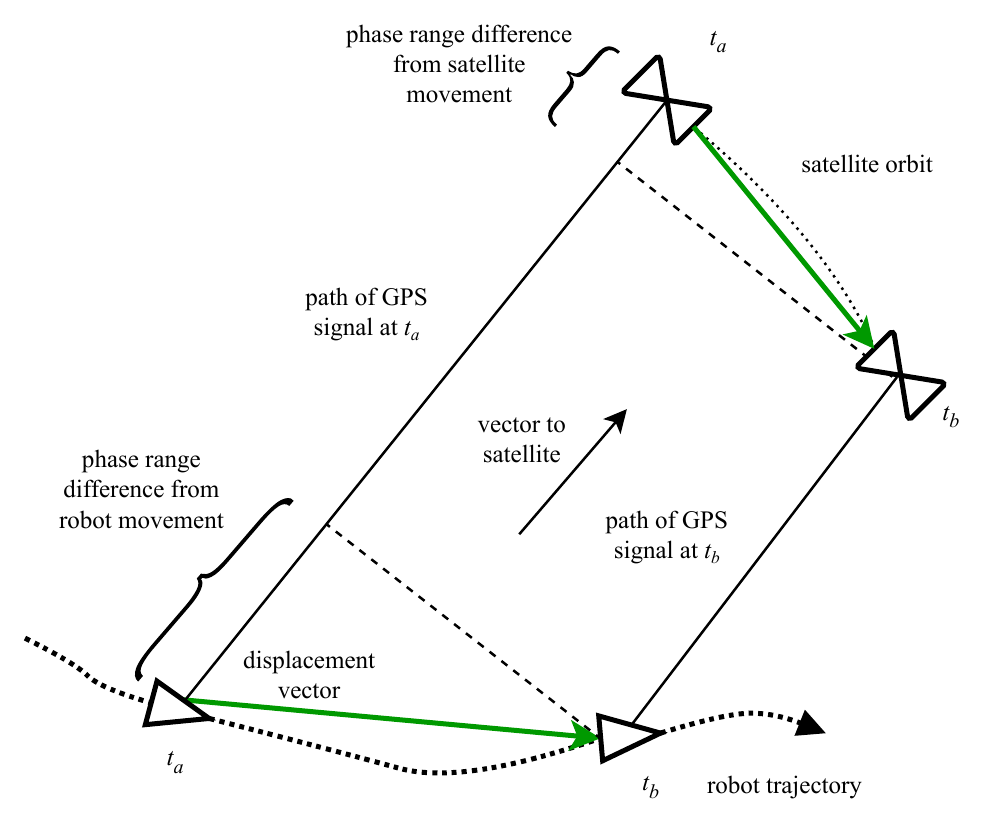}
	\caption{\footnotesize In linearizing the error term, we make the assumption that the unit vectors from receiver to satellite at times $a$ and $b$ are parallel. As a result, the difference in measured range due to receiver movement is equal to the scalar projection of the receiver displacement vector onto the satellite vector.}
	\label{fig:unit-vector-projection}
\end{figure}

For optimization, a linearized error term is needed and we derive this by noting that $\mbf{r}^{sr}_g(t_{a})$ and $\mbf{r}^{sr}_g(t_{b})$ are approximately parallel for small $t_{ba}$.
The range to the satellite can change due to both the receiver's movement and the satellite movement between measurement times.
As illustrated in Figure~\ref{fig:unit-vector-projection}, the range difference due to the receiver movement is equal to the negative of the displacement vector projected onto the unit vector to the satellite, $\hat{\mbf{u}}$:
\begin{equation}
\rho_{ba} = -\hat{\mbf{u}}^T \left( \mbf{r}^{rg}_g(t_{b}) - \mbf{r}^{rg}_g(t_{a}) \right) + \hat{\mbf{u}}^T \left( \mbf{r}^{sg}_g(t_{b}) - \mbf{r}^{sg}_g(t_{a}) \right), \label{eq:linearized-range}
\end{equation}
where the second half of the right-hand side is independent of the state.
After substituting~\eqref{eq:linearized-range} into our error equation,~\eqref{eq:tdcp_error}, we can calculate the Jacobian required to perform Gauss-Newton optimization.

\subsection{Implementation Details}\label{subsec:implementation-details}

The first step in our trajectory estimation pipeline is to parse the raw phase (logged as binary RTCM1004 messages) and calculate coarse pseudorange positions for initializing our state.
Preprocessing was done using the C library \text{RTKLIB}~\cite{takasu2009development}.
TDCP cost terms were only added between consecutive vertices in the factor graph (defined once per second) using commonly seen satellites that maintained phase lock.
This is the ``Consecutive'' factor graph configuration seen in Figure~\ref{fig:factor_graph}.
Because the majority of the error is systematic rather than the white noise, $\epsilon$, we find, as shown in Traugott~\cite{traugott2011precise}, that including TDCP constraints over longer timespans as in Suzuki~\cite{Suzuki2020} or the ``Dense'' configuration of Figure~\ref{fig:factor_graph} has no significant effect on performance besides increased computational cost.

Some of the errors in~\eqref{eq:phase-range-single}, the phase range equation, can be mitigated through modelling.
It is typical to use the Klobuchar model~\cite{klobuchar1987ionospheric} to partially correct for ionospheric effects, the parameters of which are available in the GPS navigation message.
The Niell mapping function~\cite{niell1996global} with the UNB3 model parameters~\cite{collins1996limiting} can be used to estimate the tropospheric delay.
Both models are a function of atmospheric conditions and satellite elevation.
Because atmospheric conditions change slowly and the errors are differenced in~\eqref{eq:linearized-range}, their impact is lessened compared to the effect on a single phase measurement.
However, the effect of satellite elevation change over the time difference can be significant for satellites close to the horizon.
In our experiments, we model the tropospheric delay but omit the ionospheric correction because the applicable messages were not logged for all runs.
We find the difference in performance to be negligible.

Given enough satellites, TDCP will provide a positioning solution but to be practical for vehicle odometry, and as a fair comparison for VO, we require full $SE(3)$ pose estimates in the vehicle frame.
Our algorithm is designed and tested for a nonholonomic robot so constraints that penalize lateral velocity of the vehicle frame are used to resolve vehicle orientation.
We also use a white-noise-on-acceleration (WNOA) motion prior~\cite{Anderson2015} to encourage smoothness.
Unlike other TDCP algorithms, the use of these motion models allows the robot to make use of carrier phase information and still calculate a state estimate when less than four phase-locked satellites are available.
The factor graph can be seen in Figure~\ref{fig:factor_graph}(b).

The optimization is run as a filter (forward-pass only) to simulate online odometry calculations.
It is solved with the STEAM~\cite{Anderson2015} implementation of the Dogleg Gauss-Newton algorithm~\cite{Powell1964} and the motion model applied over a 10-second sliding window.
Carrier-phase measurements are subject to outliers so a robust cost function, dynamic covariance scaling (DCS)~\cite{Agarwal2013}, is used on the TDCP factors.

\setcounter{figure}{6}
\begin{figure*}[!b]		
	\centering
	\includegraphics[trim={0 1mm 0 1mm},clip,width=0.35\textwidth]{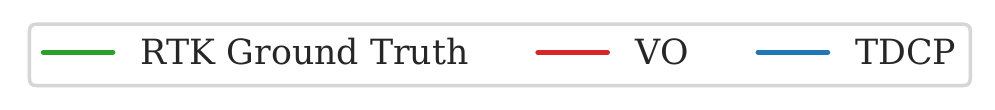} \\
	\includegraphics[width=0.32\textwidth]{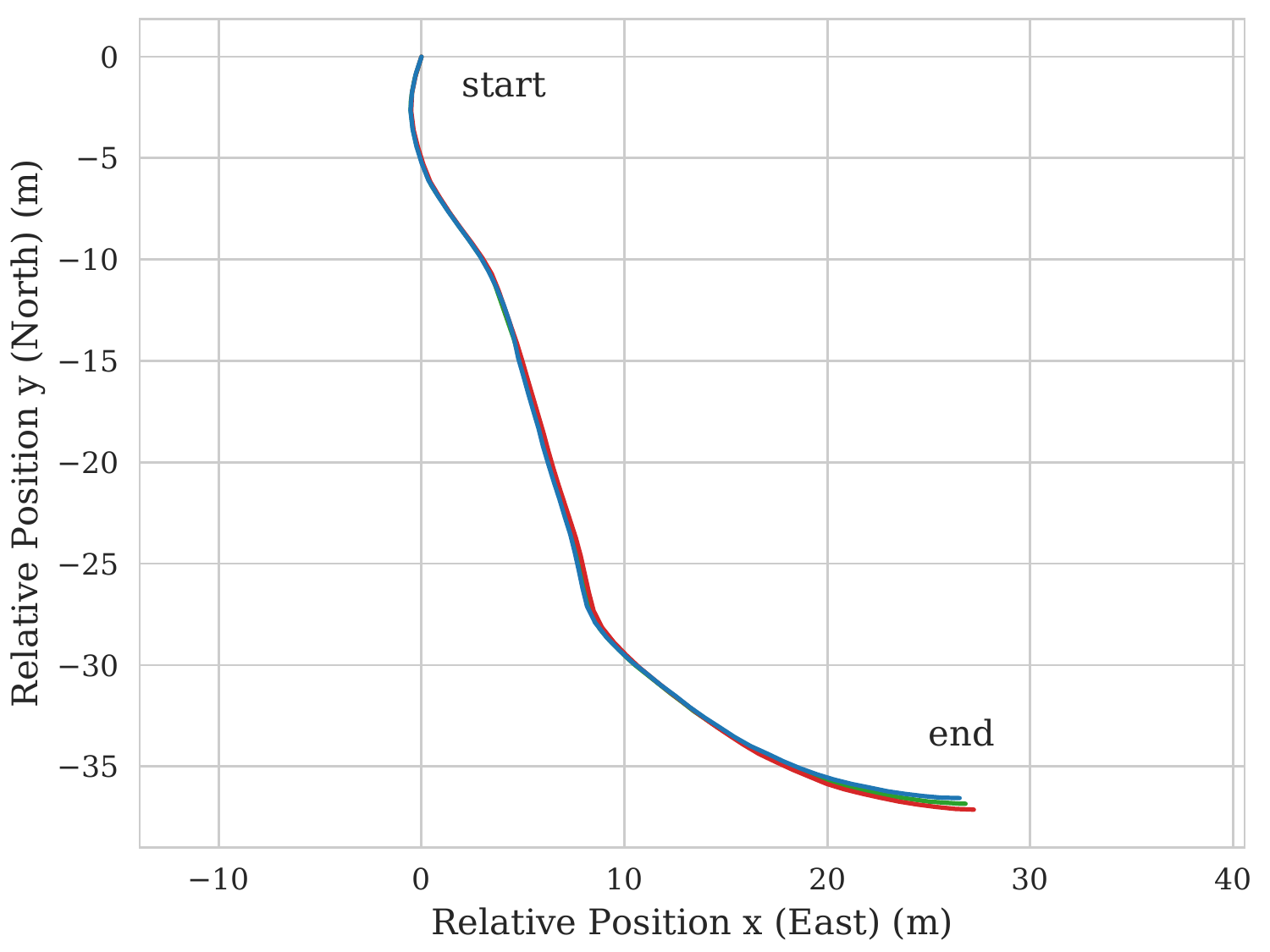}
	\includegraphics[width=0.32\textwidth]{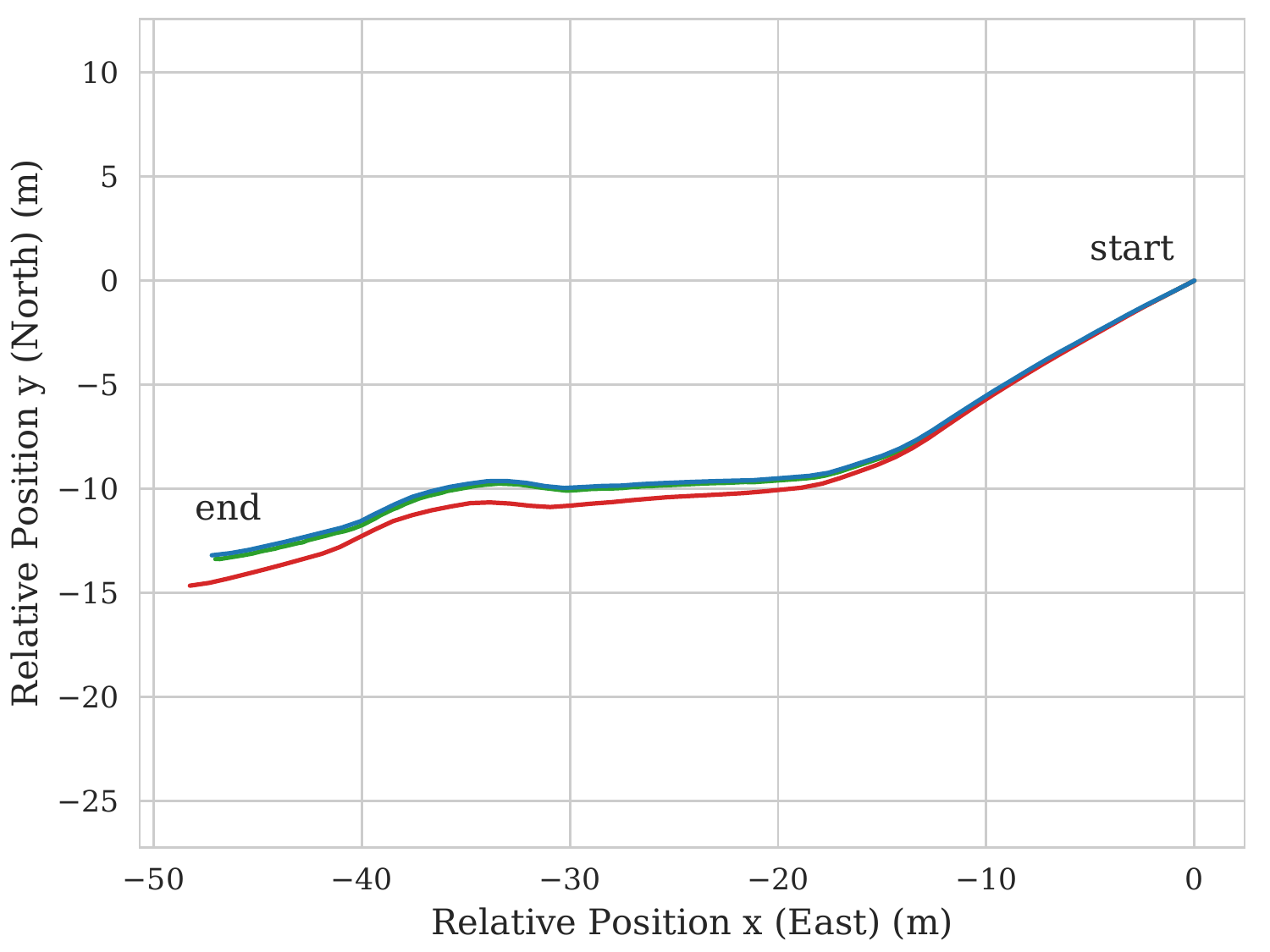}
	\includegraphics[width=0.32\textwidth]{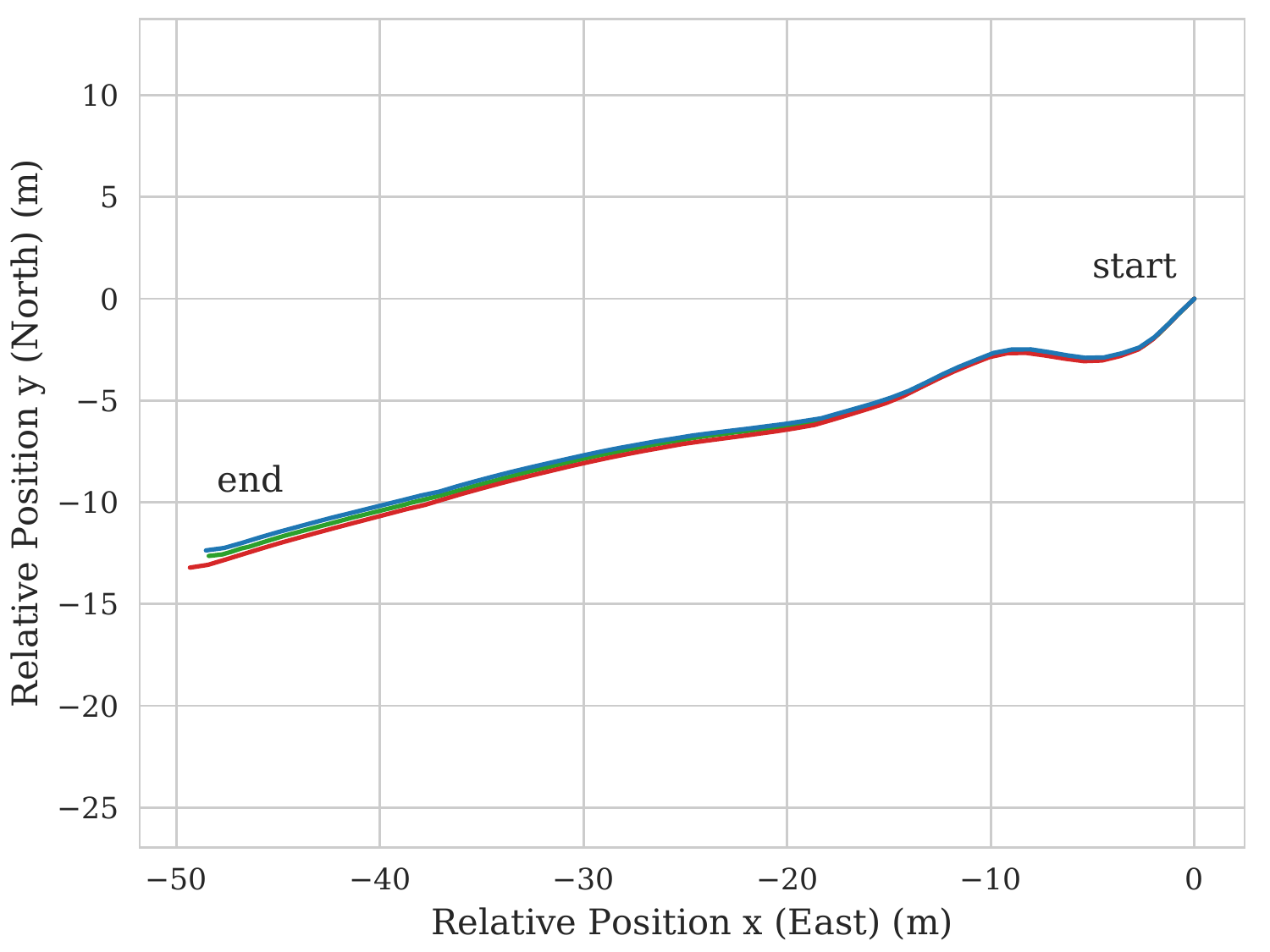}
	\caption{\footnotesize Overhead view of ground truth and estimates for three of the 15 test trajectories. VO drifts noticeably further from ground truth than the TDCP-based odometry.}
	\label{fig:test-trajectories}
\end{figure*}

\subsection{Visual Odometry}\label{subsec:visual-odometry2}

Stereo VO pose estimates are computed via the same algorithm used in VT\&R\@.
The odometry pipeline follows a similar strategy as~\cite{Klein2007} in which one module estimates camera pose with respect to the previous keyframe at framerate while another performs a local windowed bundle adjustment on map landmarks after each keyframe.
Sparse speeded-up robust features (SURF)~\cite{Bay2008} are used with random sample consensus (RANSAC)~\cite{fischler1981random} to detect outliers.
The same WNOA motion prior is used as in~\ref{subsec:implementation-details}.
The stereo error terms also have a DCS robust cost function applied to them.
Relative pose estimates are computed by solving the Gauss-Newton optimization problem with the STEAM solver.

\section{Experimental Setup}\label{sec:experimental-setup}

\setcounter{figure}{4}
\begin{figure}[tb]
	\centering
	\vspace{2mm}
	\includegraphics[width=1\linewidth]{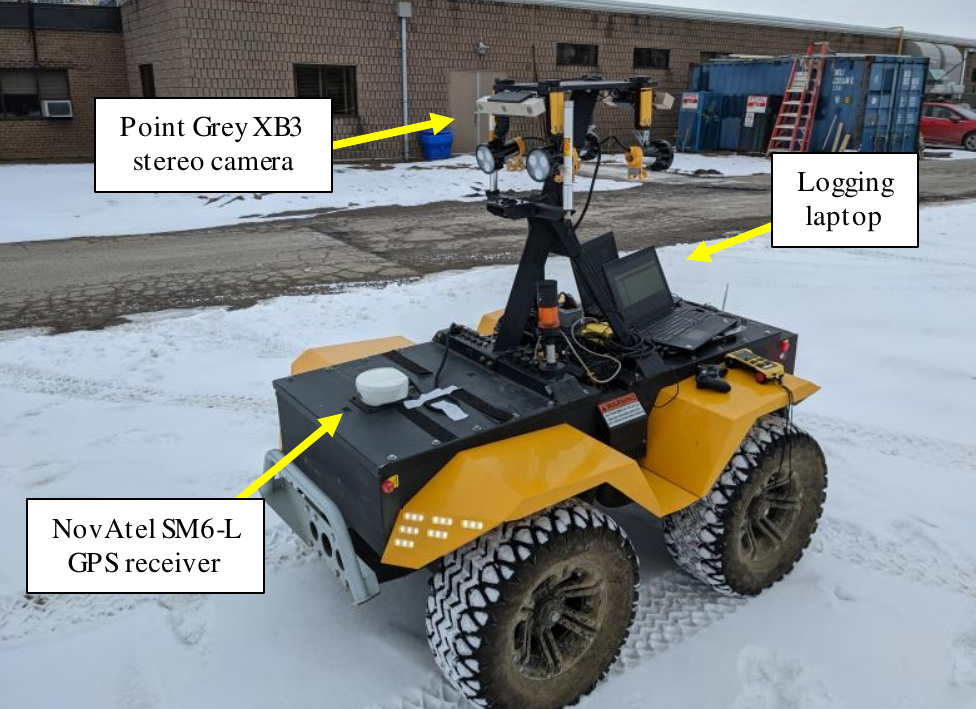}
	\caption{\footnotesize The Grizzly Robotic Utility Vehicle used for data collection.}
	\label{fig:grizzly}
\end{figure}

All data was collected aboard the Grizzly Robotic Utility Vehicle (RUV) pictured in Figure~\ref{fig:grizzly}.
The vehicle maintained an average velocity of approximately 1m/s across terrain that included pavement and snow-covered grass at the University of Toronto Institute for Aerospace Studies (UTIAS) campus.
Stereo images were captured by a front-facing Point Grey Research Bumblebee XB3 stereo camera, which has a 24cm baseline, a 66$^\circ$ horizontal field of view and captures 512x384 pixel images at a 16Hz framerate.
GPS measurements were recorded by a front-mounted NovAtel SMART6-L receiver.
Carrier phase measurements were logged at 1Hz while RTK ground truth was logged separately at 4Hz.
The RTK positioning is expected to have a RMS error of 1cm~+~1ppm under nominal conditions.
Doppler velocities were also logged for comparison.

To first test our GPS odometry, the robot was manually driven on four separate runs over two data collection days during which only GPS data was logged.
These results are presented in~\ref{subsec:gps-odometry}.
For the comparison experiment, five independent runs were driven on a third day, each spanning several minutes.
These runs were then split into 15 independent 50m sections, approximately equally spaced, for evaluation.
We chose 50m as an evaluation distance as we do not anticipate driving a robot on dead reckoning farther than this and it was sufficient for measuring odometry drift rate.
As VO does not estimate orientation in the global East-North-Up (ENU) frame, the 10m of trajectory preceding the test section was used for alignment of the VO estimates.
The continuous-time trajectories computed by STEAM are used to interpolate the VO estimates to the ground truth GPS timestamps (as they are asynchronous to the VO keyframe timestamps).
Evaluation is considered based on the amount of drift (absolute translation error) after 25m and 50m.

\section{Results}\label{sec:results}

\subsection{GPS Odometry}\label{subsec:gps-odometry}

Satellite availability for the GPS-only experiment varied throughout the runs as buildings and even the vehicle sensor mast itself caused partial occlusions of the sky.
Despite this, the receiver kept enough satellites in phase lock throughout the runs for a consistent position estimate at all times.
The median number of satellites seen was 7 with the minimum 4 and the maximum 9.

\begin{figure}[tb]
	\centering
	\vspace{2mm}
	\includegraphics[trim={0 1mm 0 1mm},clip,width=1\linewidth]{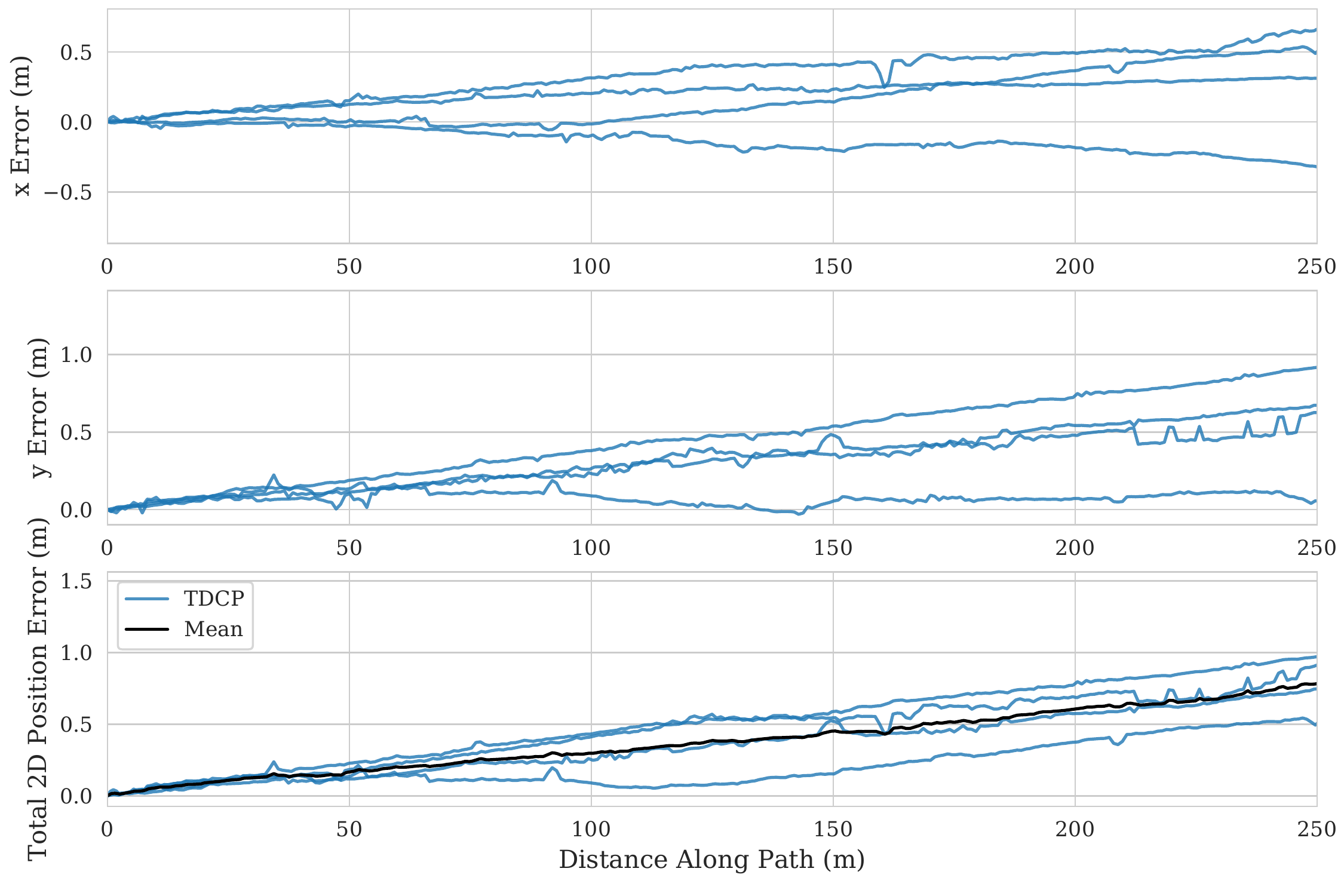}
	\caption{\footnotesize Plot of position errors over the first 250m of the four trajectories in the GPS-only experiment. The drift rate is low and the errors change approximately linearly.}
	\label{fig:gps-only-errors}
\end{figure}

Each 250m trajectory presented in Figure~\ref{fig:gps-only-errors} encompasses nearly five minutes of driving -- significantly further than a robot would need to rely on dead-reckoning between localizations.
We find that the total horizontal translational error after 250m is less than 1m for all runs and the mean error at this point is 0.78m.
The errors grow smoothly and approximately linearly.
The drift in both the $x$ (East) and $y$ (North) directions is reasonably consistent as we might expect considering the systematic errors affecting the phase measurements in Equation~\eqref{eq:phase-range-single}.
We also find the positioning errors from integrating the Doppler measurements are approximately twice as large as in our TDCP algorithm.
The results provide further evidence that TDCP should be preferred over integrated Doppler velocity for estimating relative position with a single receiver.

\setcounter{figure}{8}
 \begin{figure*}[!b]
	\centering
	\includegraphics[trim={0 1mm 0 1mm},clip,width=0.75\textwidth]{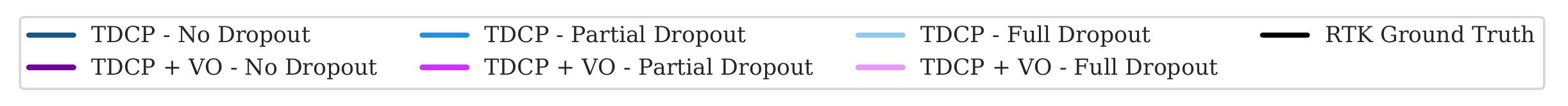}
	\includegraphics[trim={0 1mm 0 1mm},clip,width=0.245\textwidth]{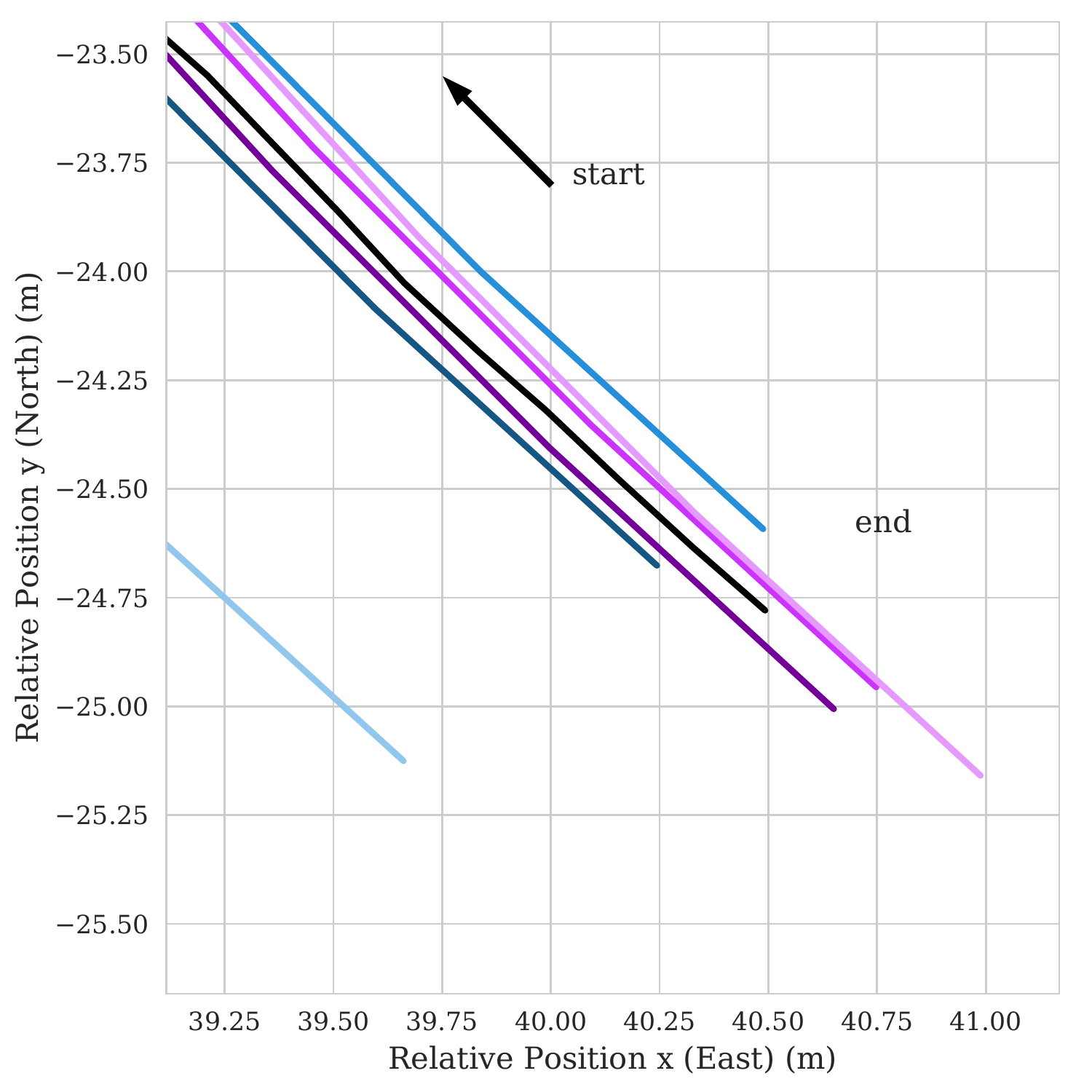}
	\includegraphics[trim={0 1mm 0 1mm},clip,width=0.245\textwidth]{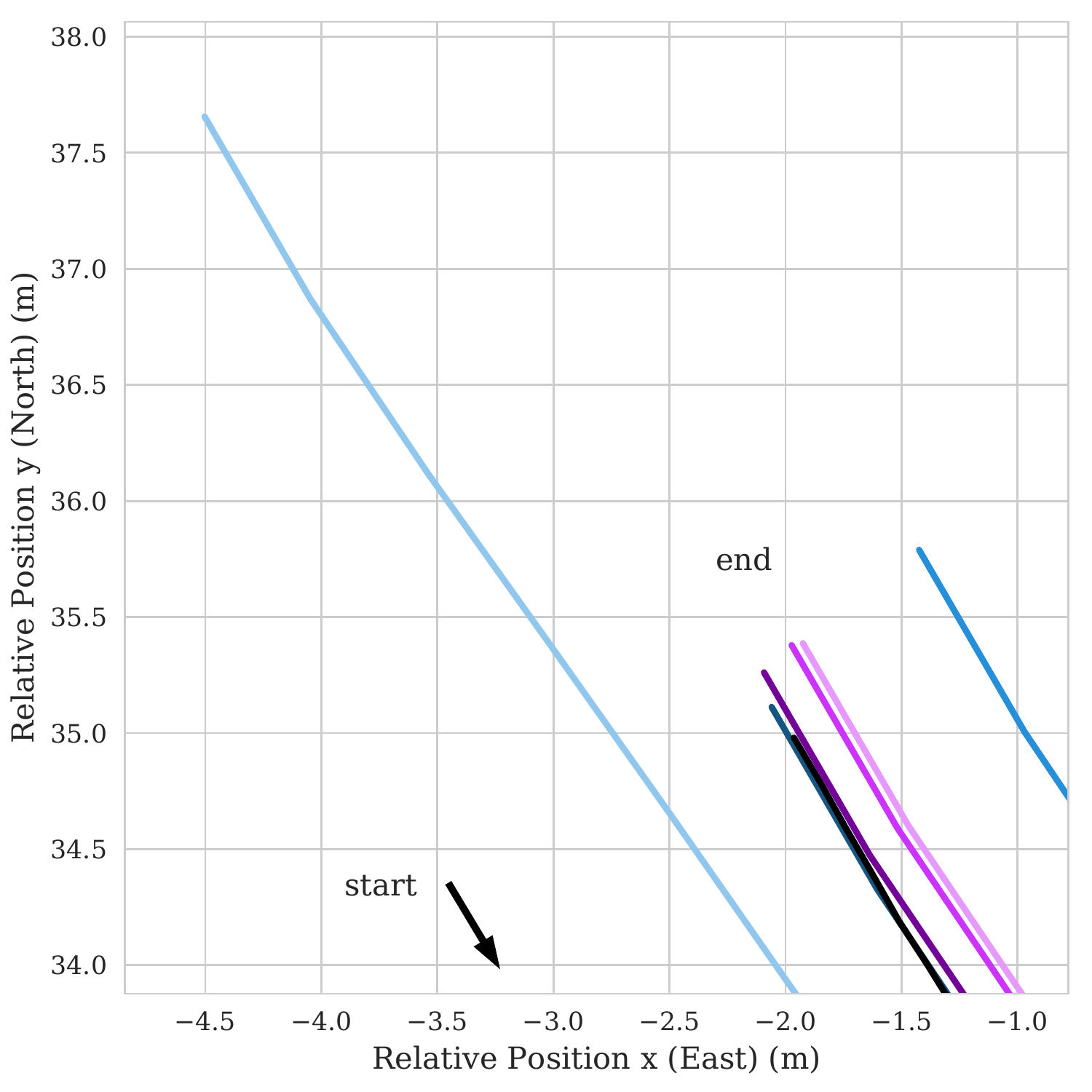}
	\includegraphics[trim={0 1mm 0 1mm},clip,width=0.245\textwidth]{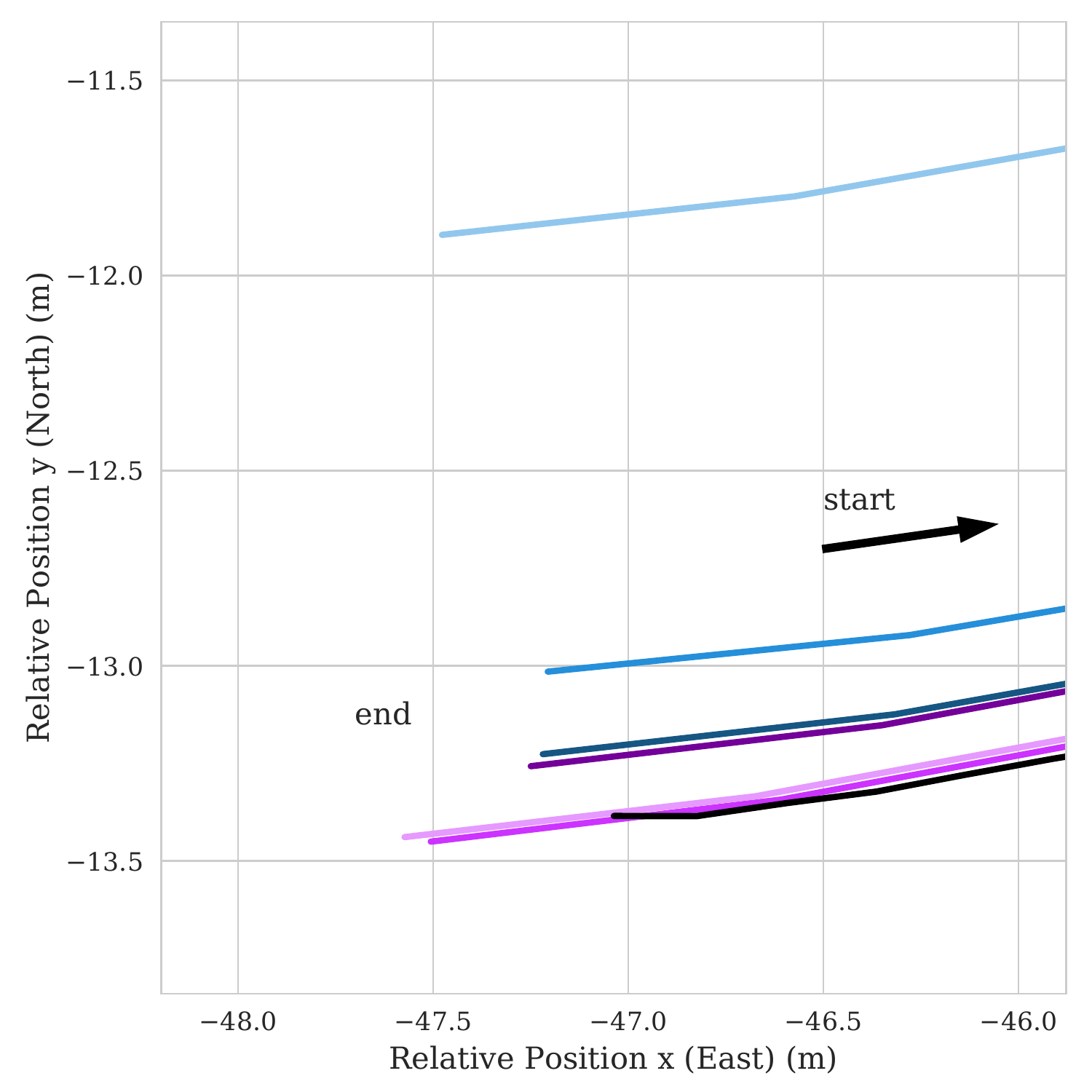}
	\includegraphics[trim={0 1mm 0 1mm},clip,width=0.245\textwidth]{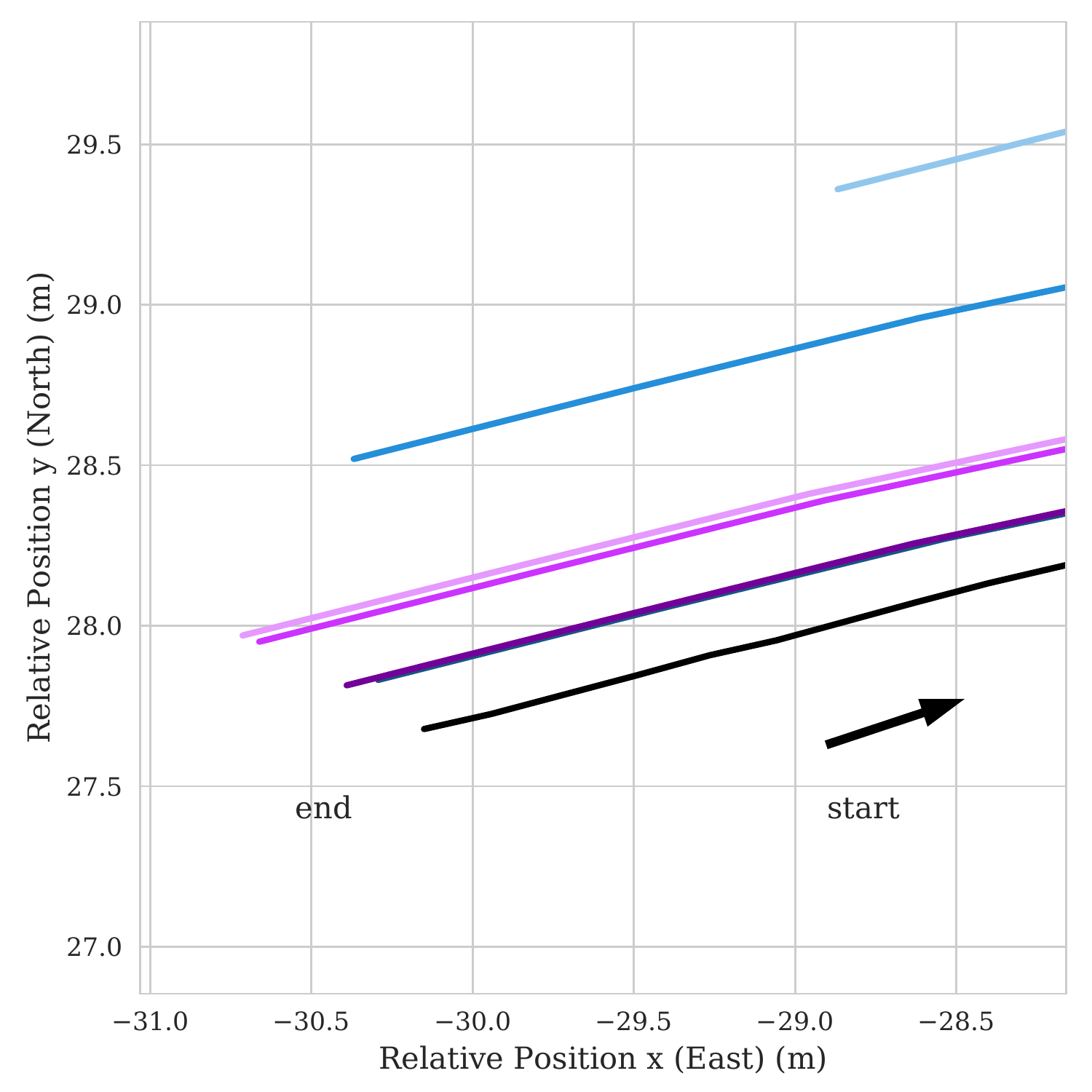}
	\caption{\footnotesize Plot of final positional error for four 50m paths illustrating the effect of GPS dropout on our TDCP algorithm. The addition of the lower quality VO measurements to our algorithm makes little difference when enough satellites are available. However, the TDCP+VO algorithm is much more robust when a short satellite dropout is simulated in the trajectory.}
	\label{fig:dropout}
\end{figure*}

\subsection{Comparison to Visual Odometry}\label{subsec:comparison-to-visual-odometry}     

Figure~\ref{fig:test-trajectories} shows an overhead view of the estimates from both algorithms on three of the test trajectories representative of the larger test set.
Even at this macroscopic scale we can see the GPS odometry outperforms VO\@.
Figure~\ref{fig:vo-gps-errors} depicts both the errors for the individual runs and an average horizontal position error for each algorithm.
After 50m, the TDCP method has a smaller translational error than VO on all but one of the 15 test trajectories.
VO has a mean final translational error of 1.127m or 2.25\% while TDCP does 75\% better with a mean error of 0.281m or 0.56\%.
The results are similar after just 25m, with drift rates of 2.26\% and 0.57\%, respectively.
The variance in drift rate between runs is also a lot higher for VO as can be seen in the spread of data in Figure~\ref{fig:vo-gps-errors}.
This implies the expected errors may be more predictable for TDCP\@.

\setcounter{figure}{7}
\begin{figure}[tb]
	\centering
	\vspace{2mm}
	\includegraphics[width=1\linewidth]{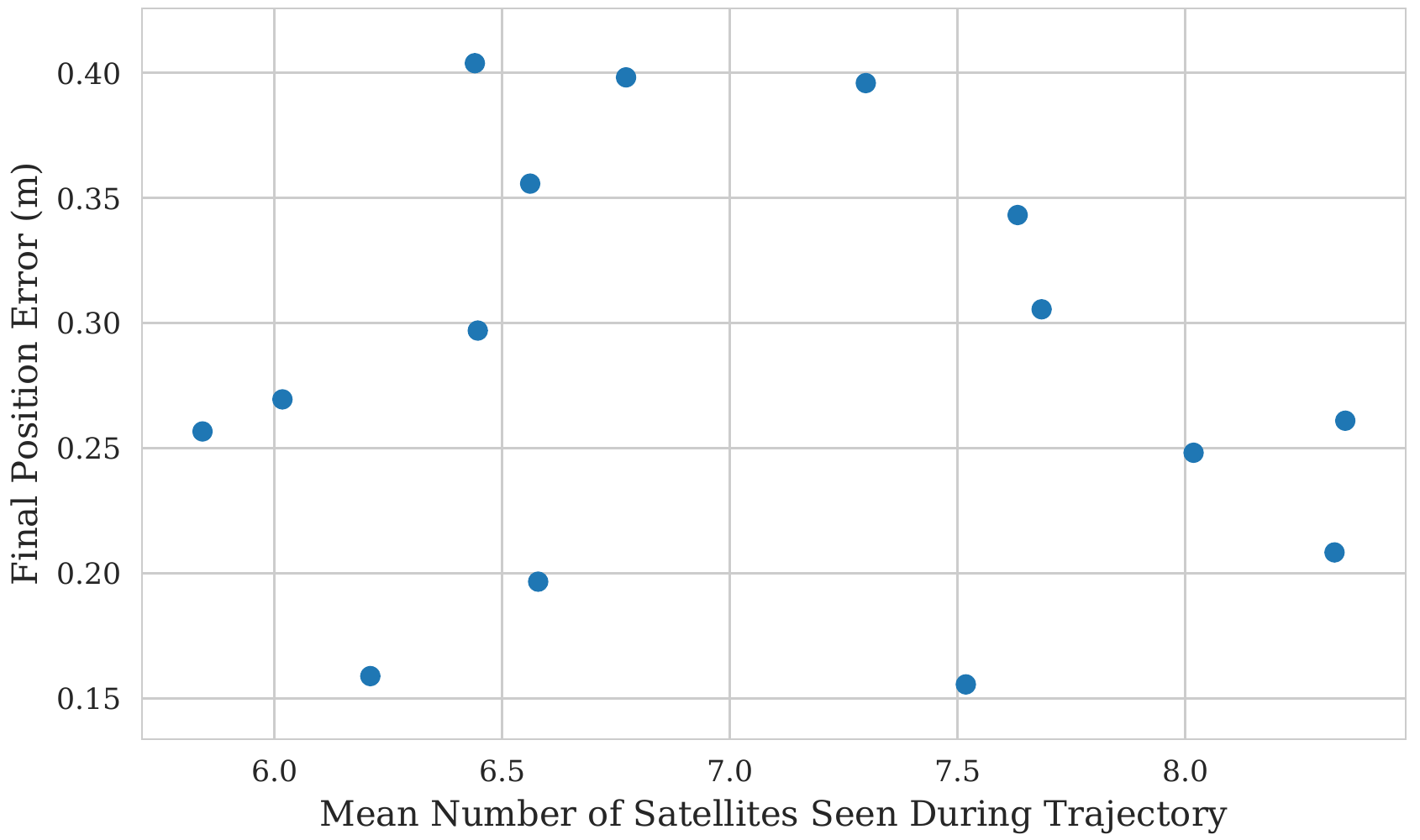}
	\caption{\footnotesize Scatter plot of the final position error for TDCP across the test trajectories versus the mean number of satellites observed. There is a weak negative correlation suggesting having more satellites improves accuracy but other factors are also important.}
	\label{fig:sat-scatter}
\end{figure}

A similar number of satellites were available for the comparison experiment as in~\ref{subsec:gps-odometry} with the minimum 5, the median 7, and the maximum 9.
Figure~\ref{fig:sat-scatter} examines the relationship between mean satellites seen over a trajectory and final error.
There is a negative correlation between errors and number of satellites as expected though the relationship is weak ($r=-0.10$).
Other factors such as the particular geometry of the satellites, the atmospheric conditions, and the shape of the trajectory may have more influence.

Looking more closely at the VO results, we see the number of feature matches varies somewhat between the two major types of terrain seen -- dry pavement and snow, but is enough for a reasonable motion estimate throughout.
There were no VO failures (i.e., there were always enough landmark matches to produce a consistent estimate).
We notice the VO tends to slightly overestimate or underestimate distances within a run.
As a result the total translational error over a full loop trajectory is smaller than the drift rate for shorter sections though still significant.
Further tuning might be able to improve VO performance slightly, but it is unlikely to reach the level of the GPS odometry.
Finally, we note the TDCP method also has computational advantages as it only requires one error term per satellite pair compared to the potentially hundreds of stereo landmark terms involved in VO.
In our head-to-head comparison, GPS was clearly superior.

\subsection{Combining GPS and Vision}\label{subsec:combining}

As our GPS odometry algorithm has been set up as a factor graph, it is amendable to adding factors from other sensors.
A natural choice would be to combine the visual and GPS odometry estimators as shown in Figure~\ref{fig:factor_graph} (c) and (d).
The results in this section are from a loosely coupled estimator for ease of comparison.
We find under good conditions the addition of vision does not significantly improve accuracy because the estimates from VO are worse quality.
If the uncertainties are improperly set, the inclusion can actually degrade performance.
But, using both sensors does improve robustness when the quality or availability of one or both sensors cannot be guaranteed.
To show this, we simulate both full (zero satellites available) and partial (two satellites available) temporary GPS dropouts and observe the effect on our odometry with and without the inclusion of VO.
The short, 15-second dropouts occur near the beginning of the approximately one-minute long trajectories.

As seen in Figure~\ref{fig:dropout}, qualitatively, we get similar estimates with and without VO when sufficient satellites are available throughout.
During dropouts, the GPS-only estimator is forced to rely heavily on its motion model and accuracy suffers.
Local accuracy does recover once satellites are reacquired.
In the partial dropout, the receiver displacement is not fully constrained as only two satellites are available, but our algorithm can still make use of the carrier phase information to some degree and performance is much better than with zero satellites.
However, in many applications, the added error would still be considered a failure.
With the addition of VO, the performance loss from the dropout is barely noticeable.
A combined approach provides the added accuracy of TDCP with the reliability of VO\@.

\section{Conclusions and Future Work}\label{sec:conclusions-and-future-work}

\addtolength{\textheight}{-4cm}   

We simultaneously collected a large set of GPS data and stereo imagery from a ground robot driving outdoors.
We evaluated our TDCP-based single-receiver, single-frequency GPS odometry algorithm against a proven stereo VO pipeline in the first known experiment of this kind.
The results showed the GPS odometry produced far smaller positional errors with respect to the RTK ground truth.
We believe TDCP odometry is a good alternative to VO for outdoor navigation.
VO is still preferred in areas where occlusions or other sources of GNSS signal interference are a frequent issue.
For added robustness, or in applications such as indoor-outdoor navigation, the two sensors may be combined.

One likely way to improve the future performance of our positioning algorithm is to incorporate additional GNSS constellations thereby increasing the number of satellites available.
Because the error drift in our estimates is approximately constant throughout a run, it may be possible to estimate and correct for this bias using other sensors.
Incorporating TDCP estimates into multi-experience localization (MEL)~\cite{Paton2016} is a final opportunity for future work of particular interest to us.
MEL can fail due to high appearance change.
A better odometry solution would allow the robot to safely continue autonomous path traversal while simultaneously logging images to update the map and improve future chances of successful visual localization.


\section*{Acknowledgment}
We would like to thank the Natural Sciences and Engineering Research Council of Canada (NSERC) and Clearpath Robotics for supporting this work.

\bibliographystyle{IEEEtran}
\bibliography{bib/refs}

\end{document}